\documentclass{article}
\usepackage{ijcai24}

\usepackage{times}
\usepackage{soul}
\usepackage{url}
\usepackage[hidelinks]{hyperref}
\usepackage[utf8]{inputenc}
\usepackage[small]{caption}
\usepackage{graphicx}
\usepackage{amsmath}
\usepackage{amsthm}
\usepackage{booktabs}
\usepackage{algorithm}
\usepackage{algorithmic}
\usepackage[switch]{lineno}
\usepackage{pifont}
\usepackage{enumitem}
\usepackage{verbatim}
\usepackage{makecell}
\usepackage{xcolor}
\usepackage{hyperref}
\usepackage{url}
\usepackage{multirow}
\usepackage{booktabs}
\usepackage{graphicx}
\usepackage{dsfont}
\usepackage{wrapfig}
\usepackage{subfig}
\usepackage{overpic}
\usepackage{color}
\definecolor{mydarkblue}{RGB}{0, 0, 255}
\usepackage{times}
\usepackage{epsfig}
\usepackage{amsthm, amsmath}
\usepackage{mathrsfs,amssymb}
\usepackage{pifont}
\usepackage{enumitem}
\usepackage{verbatim}
\usepackage{makecell}
\usepackage{xcolor}
\usepackage{multirow}
\usepackage{utfsym}
\usepackage{tabularray}

\usepackage{algorithm}
\usepackage{algorithmic}

\title{Chain-of-Memory: Enhancing GUI Agents for Cross-Application Navigation}

\author{
Xinzge Gao$^{1,2}$
\and
Chuanrui Hu$^2$\thanks{Project Leader.}\and
Bin Chen$^2$\And
Teng Li$^1$\thanks{Corresponding author.}\\
\affiliations
$^1${Anhui University}\\
$^2${Qihoo360}\\
\emails
\{gaoxingze, huchuanrui, chenbin\}@360.cn,
tenglwy@gmail.com
}
\begin{document}

\maketitle

\begin{abstract}
    Multimodal large language models (MLLMs) are attracting growing attention in the development of Graphical User Interface (GUI) agents. Existing approaches often rely on historical screenshots or actions to implicitly represent the task state. This reliance poses challenges for GUI agents in accurately understanding task states and underscores the absence of effective mechanisms to store critical information in complex and lengthy cross-app tasks. To address these challenges, we propose Chain-of-Memory (CoM), a novel approach for explicitly modeling short-term and long-term memory in GUI agents. CoM achieves this by capturing action descriptions, integrating task-relevant screen information, and maintaining a dedicated memory module to store and manage this information. By leveraging explicit memory representations, CoM enables GUI agents to better understand task states and retain critical historical information persistently. To equip GUI agents with memory management capabilities and evaluate the effectiveness of CoM, we developed the GUI Odyssey-CoM, a dataset comprising 111k screen-action pairs annotated with Chain-of-Memory. Experimental results demonstrate that CoM significantly improves GUI agents' performance in cross-application tasks. Additionally, GUI Odyssey-CoM enables 7B models to achieve memory management capabilities comparable to 72B models. The dataset and code will be open-sourced.
\end{abstract}

% \begin{figure*}[H]
% \vspace{-0.3cm}
% \setlength{\belowcaptionskip}{0cm}
% \setlength{\abovecaptionskip}{0.1cm}
%     \centering
%     \includegraphics[width=\linewidth]{fig/sample.pdf}
%     \caption{\textbf{Each component and workflow of Chain-of-Memory (CoM):}(a)The agent first updates its current memory using the current screen and previous information, then makes decisions based on the current screen and memory information.(b) The history length of a general GUI Agent is limited, leading to the forgetting of previously viewed content.(c) Relying solely on the memory of the previous action is insufficient to fully grasp the progress of a task.}
%     \label{fig:sample}
%     \vspace{-0.2cm}
% \end{figure*}

\section{Introduction}
\label{sec:intro}

\begin{figure}[t!]
  \centering %居中
  \includegraphics[width=1.0\linewidth]{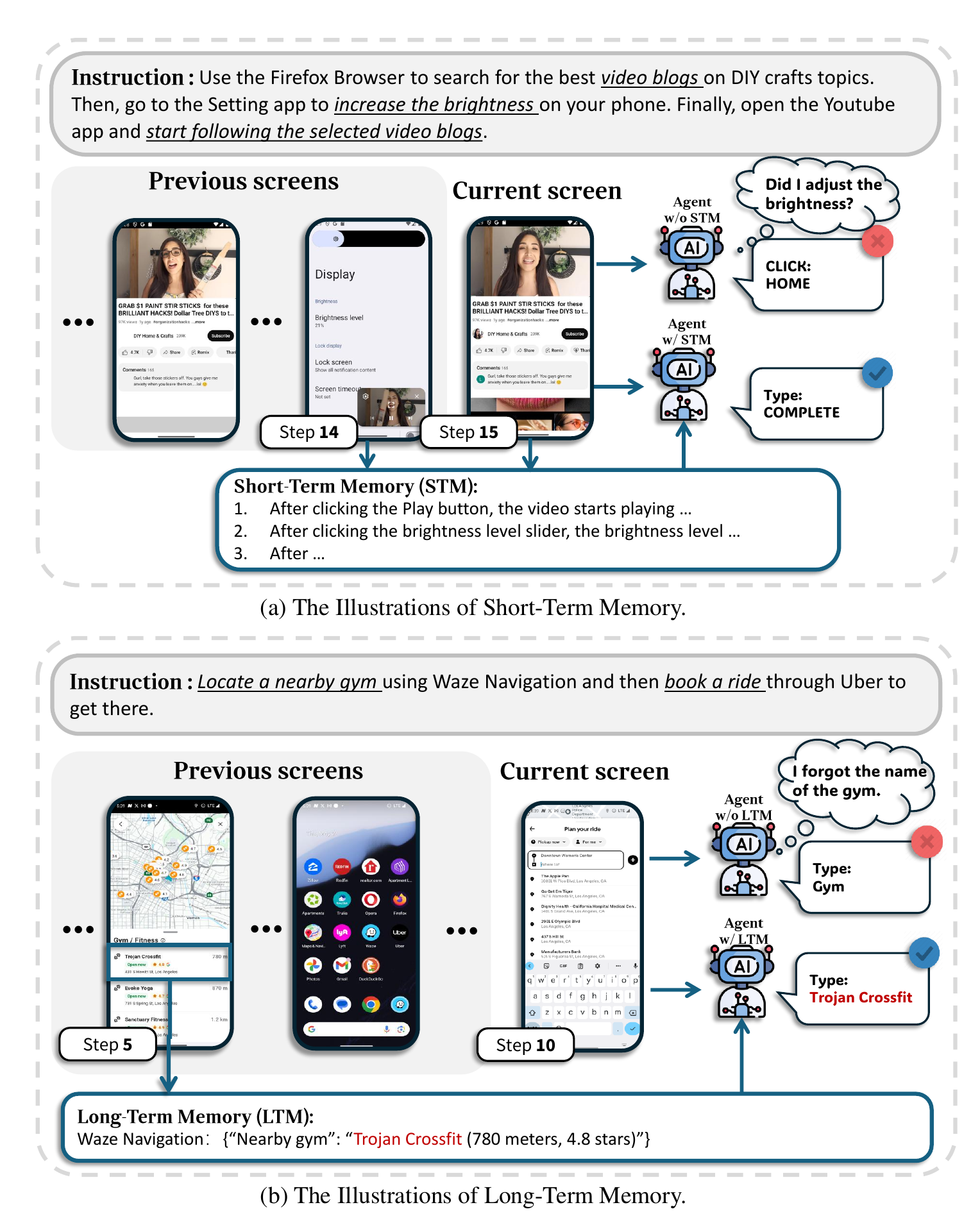}
  \caption{(a) Previous methods that only utilized action history as memory information were insufficient for the agent to understand the task state. The proposed short-term memory in the form of text descriptions helps the agent comprehend the task state. (b) Previously, there was a lack of means to retain distant memories. We propose a method that can extract and long-term preserve memory.}
  \label{fig:sample} %指定给图片一个标签
\end{figure}

\begin{table*}[]
\setlength{\abovecaptionskip}{0.15cm}
\setlength{\belowcaptionskip}{0cm}
\centering
\resizebox{160mm}{!}{
\begin{tabular}{l|cccc|c|c|c|c|c|c}
\toprule
\multirow{3}{*}{\textbf{Dataset}}   & \multirow{3}{*}{\textbf{\#Episodes}} & \multirow{3}{*}{\begin{tabular}[c]{@{}c@{}}\textbf{\#Unique}\\ \textbf{Instructions}\end{tabular}} & \multirow{3}{*}{\textbf{\#Apps}} & \multirow{3}{*}{\textbf{\#Steps}} & \multirow{3}{*}{\textbf{Cross-app}} & \multicolumn{5}{c}{\textbf{Annotation}}   \\ \cmidrule{7-11}
  & & & & & & \begin{tabular}[c]{@{}c@{}}\textbf{screen}\\ \textbf{info}\end{tabular} & \begin{tabular}[c]{@{}c@{}}\textbf{action}\\ \textbf{coord}\end{tabular} & \begin{tabular}[c]{@{}c@{}}\textbf{action}\\ \textbf{desc}\end{tabular} & \begin{tabular}[c]{@{}c@{}}\textbf{action}\\ \textbf{thinking}\end{tabular} & \begin{tabular}[c]{@{}c@{}}\textbf{Long-term}\\ \textbf{Memory}\end{tabular} \\ \midrule
PixelHelp~\cite{li2020mapping}      & 187     &  187   &  4    & \textasciitilde 4  &             &              & $\checkmark$    &               &               &   \\
MoTIF~\cite{burns2021mobile}       & 4707    &  270   & 125   & 4.5                &             &              & $\checkmark$    &               &               &  \\
UGIF~\cite{venkatesh2022ugif}       & 523     &  480   & 12    & 6.3                &             &              & $\checkmark$    &$\checkmark$   &               &   \\
Meta-GUI~\cite{sun2022meta}    & 4684    & 1125   & 11    & 5.3                &             &              & $\checkmark$    &               &               &   \\
AITW~\cite{rawles2023android}       & 715142  & 30378  & 159   & 6.5                &             &              & $\checkmark$    &               &               &   \\ 
Mind2Web~\cite{deng2024mind2web}   & 2350    & 2350   & 137   & 7.3                &             &              & $\checkmark$    &               &               &   \\ 
AITZ~\cite{zhang2024android}                                 & 2504    & 2504   & 70+   & 7.5                &             & $\checkmark$ & $\checkmark$    & $\checkmark$  & $\checkmark$  &  \\
GUI Odyssey~\cite{lu2024gui}                          & 7735    & 7735   & 201   & 15.4               & $\checkmark$ &              & $\checkmark$    &               &               &              \\ \midrule
\textbf{\textsc{GUI Odyssey-CoM}}               & 7735    & 7735   & 201   & 15.4               & $\checkmark$ & $\checkmark$ & $\checkmark$    & $\checkmark$  &               & $\checkmark$ \\
\bottomrule
\end{tabular}}
\caption{\textbf{Comparison with previous datasets.} GUI Odyssey-CoM is re-annotated and utilized based on the GUI Odyssey.}
\end{table*}

GUI Agents are primarily designed to automate the execution of complex tasks involving multiple applications on GUI-based devices. By interpreting and processing screen information along with user instructions, these agents facilitate seamless navigation across diverse applications, thereby enhancing user experience. Previous work utilized the GPT-4 series models with prompt engineering to develop a GUI Navigation Agent~\cite{zhang2023appagent,li2024appagent,ufo,wang2024mobile,wang2024mobilev2}. However, relying solely on prompt engineering failed to address the absence of GUI-related knowledge and recognition capabilities. Consequently, subsequent work focused on developing GUI Navigation Agents by fine-tuning multimodal large models using GUI-specific datasets~\cite{rawles2023android,venkatesh2022ugif,li2024effects,lu2024gui,baechler2024screenai,you2025ferret,wan2024omniparser}.

Despite the improvements, most of these approaches still depend on historical actions (e.g., Click (x,y)) or  purely visual historical screenshots to infer the current task state and make subsequent decisions. Historical actions provide very limited information, and historical screens contain redundant information with only a small fraction being relevant to the task state. Therefore, these methods struggle to accurately reflect the current task state. Additionally, the context window size of multimodal large language models (MLLMs) limits the retention of task information from historical screens, leading to the loss of critical information (e.g., prior search results) after a certain number of steps, thus posing a challenge for the agent's decision-making process.

Inspired by the similarity between the process of information exchange between the decision-making center and storage system during human work~\cite{baddeley2003working} and the information management mechanism required by GUI Agent, we propose a novel paradigm named Chain-of-Memory (CoM). Specifically, like humans navigating complex tasks, GUI agents require mechanisms to retain short-term task context, access relevant long-term information across applications, and filter out irrelevant screen details. 

CoM comprises two essential components: \textbf{Short-Term Memory (STM)} and \textbf{Long-Term Memory (LTM)}. \textbf{STM} is tasked with storing language descriptions of recent operations, ensuring the agent understands the most recent task context to make informed decisions. \textbf{LTM} is responsible for storing information that may be required in the future, such as search results or task-specific knowledge. To facilitate the functionality of LTM, CoM incorporates \textbf{Screen Information} component. This component is designed to capture task-relevant key information displayed on the screen and filter out irrelevant elements. The extracted information then serves as the data source for LTM.

\begin{comment}
CoM incorporates three key components:
\begin{itemize}
    \item \textbf{Short-Term Memory (STM)}: STM stores real-time information about the immediate task state, ensuring the agent has access to the most recent context.
    \item \textbf{Screen Information}: This component captures only task-relevant information from the screen, filtering out irrelevant UI elements and background data to focus on critical content.
    \item \textbf{Long-Term Memory (LTM)}: LTM stores essential screen information, such as user preferences or task-specific knowledge, that may be needed across different stages of a task or even in future tasks.
\end{itemize}
\end{comment}

Based on the components described above, the CoM pipeline is organized into four steps: \textbf{(1) Information Perception:} extract and summarize data from previous screenshots and historical operations. \textbf{(2) STM Update}: integrate the recently acquired information into short-term memory. \textbf{(3) LTM Storage}: identify and store the critical information from previously collected data into long-term memory. \textbf{(4) Action Decision:} evaluate the current task status by leveraging STM and LTM, and make subsequent decisions accordingly.

Generation and utilization of memory information require strong reasoning capabilities, but previous studies~\cite{kojima2022large,wei2022chain} have shown that only large-scale models possess such capabilities, while smaller models struggle in this regard. To empower smaller models with memory generation and utilization capabilities through instruction fine-tuning, we introduce the \textbf{GUI Odyssey-CoM}, a pioneering dataset that uniquely integrates memory processes with action decision-making. It encompasses a substantial collection of 111,791 screen-action pairs, each annotated with relevant memory information. Leveraging the GUI Odyssey-CoM, we conducted zero-shot and fine-tuning evaluations. The experimental results demonstrate that CoM significantly improves operational accuracy, task success rates, and task switching success rates in cross-application tasks. Furthermore, fine-tuning enabled the 7B model to effectively utilize memory information, achieving comparable memory information generation capabilities to the 72B model, thereby validating the effectiveness of CoM and underscoring the necessity of the GUI Odyssey-CoM.

The contributions of this work are as follows:
\begin{itemize}
    \item We propose Chain-of-Memory (CoM), a novel method that enables a GUI Agent to recognize and memorize tasks in a human-like manner. This approach allows the Agent to better understand the current task status and effectively handle multi-step tasks across different applications.
    \item We have constructed the GUI Odyssey-CoM dataset, currently the largest cross-app GUI Navigation dataset that encompasses multiple high-quality text information annotation.
    \item We conducted zero-shot and fine-tuning evaluations using the GUI Odyssey-CoM. These evaluations demonstrate the effectiveness of CoM in enhancing the performance of GUI agents and highlight the necessity of the GUI Odyssey-CoM for enabling smaller models to develop memory management capabilities.
\end{itemize}

\begin{figure*}[t]
\vspace{-0.3cm}
\setlength{\belowcaptionskip}{0cm}
\setlength{\abovecaptionskip}{0.1cm}
    \centering
    \includegraphics[width=\linewidth]{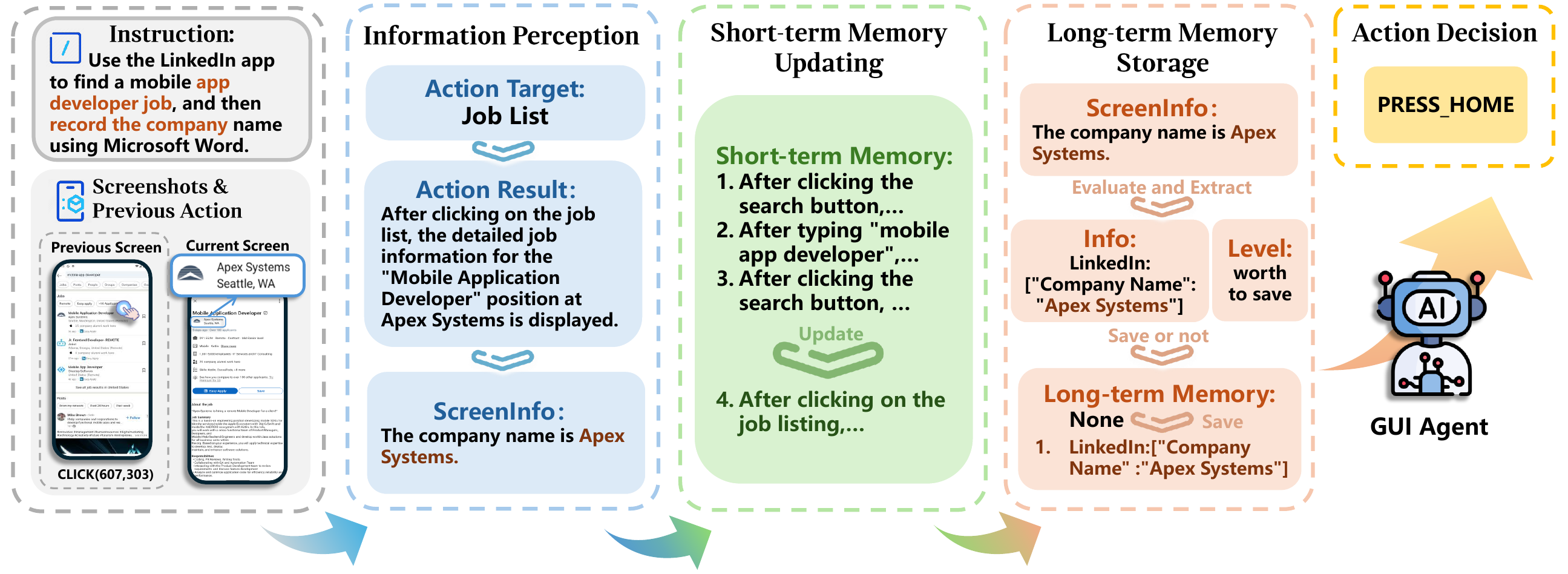}
    \caption{\textbf{The architecture and workflow of Chain-of-Memory.} CoM comprises four key stages: Information Perception, STM Updating, LTM Storage, and Action Decision. The agent perceives information from previous and current screens, updates the STM with recent action results, stores key information in the LTM, and makes the next action decision based on the current screen, STM, and LTM.}
    \label{fig:overall}
    \vspace{-0.2cm}
\end{figure*}
\section{Related work}
\label{sec:related}

\subsection{MLLMs as GUI Agent}
Driven by advancements in areas such as multimodal fusion~\cite{radford2021learning,zhang2023llama,koh2023grounding,team2024chameleon} and dynamic resolution~\cite{wang2024qwen2,chen2024internvl}, MLLMs are progressively demonstrating the potential to function as effective Agents. Employing LLM and MLLM as agents to assist human beings has emerged as a research focus, such as Code Agent~\cite{liu2024empirical,qian2023communicative}, Gaming Agent~\cite{meta2022human}, and GUI Agent~\cite{ufo,zhang2023appagent,wang2024mobile,wang2024mobilev2,cheng2024seeclick}. UFO~\cite{ufo} has designed an dual Agent structure. It utilizes the APIs of the Windows system to extract UI information and completes operations on Windows through the obtained UI information and screenshots. AppAgent~\cite{zhang2023appagent} utilizes Android tools to extract UI information into XML files and has designed an Agent that performs operations on mobile apps by using screenshots and the UI information in the XML files. To eliminate the dependency on system APIs, Mobile-Agent~\cite{wang2024mobilev2} has designed a visual multi-agent collaborative system that executes human commands using pre-trained MLLMs. Constrained by the limitations of training data, the GUI agents designed with pre-trained MLLMs exhibit significant limitations in their application. SeeClick~\cite{cheng2024seeclick} has created the first real GUI basic dataset named ScreenSpot that covers mobile, desktop, and web environments, and has trained a GUI Agent with a broader range of applications using this data.

\subsection{GUI Navigation}
Previously developed GUI Agents constructed with pre-trained models typically rely on system mechanisms~\cite{ufo,zhang2023appagent} or external tools~\cite{wang2024mobile} for UI grounding. The emergence of GUI Navigation datasets, such as AITW~\cite{rawles2023android}, UGIF~\cite{venkatesh2022ugif}, GUI-World~\cite{chen2024gui}, GUI Odyssey~\cite{lu2024gui}, which contain screenshot action pairs, allows GUI-Agents to achieve independent grounding and operate UIs in more tasks.Subsequently, AITZ~\cite{zhang2024android} refined AITW~\cite{rawles2023android} and added a detailed reasoning process. This allows the model to learn the true significance and context relationship of Action. However, the instructions in previous datasets generally only contain simple tasks for a single application. To better fulfill real-world requirements, GUI Odyssey~\cite{lu2024gui} has released a multi-step, multi-device, and cross-application GUI Navigation dataset. As tasks grow progressively more complex, relying only on historical action information is hardly sufficient to meet the requirements for Agent decision. Our CoM mitigates this issue.

%-------------------------------------------------------------------------

\section{The Proposed Method}
This section details the workflow and individual CoM modules. As illustrated in Figure \ref{fig:overall}, the CoM framework operates in a sequential manner, encompassing information perception, short-term memory updating, long-term memory storage, and next-step decision-making. During the information perception phase, the GUI Agent monitors changes in the operational area of the screen and extracts critical information from the current screen. The perceived information is then used to update the short-term memory, while essential details are stored in the long-term memory. Finally, leveraging these accumulated insights, the state of the task is assessed and the appropriate next action is determined.

%从认知过程的流程来看，人类是按照「信息感知→注意过滤→短期记忆→长期记忆存储→行为决策」的路径进行。我们仿照这种机制
\subsection{Short-term Memory}
In GUI-related tasks, decisions for subsequent actions are often highly dependent on the current task state, which is shaped by historical information. Humans process historical information by abstracting previously observed visual information and executed actions into a lower-dimensional representation. Prior approaches have leveraged historical data in the form of simple actions, such as “Click(x,y)” or entire historical screens. However, simple action logs lack sufficient detail, while raw screens introduce excessive redundancy, often obscuring critical information within irrelevant data.

To address this issue, we compare screen states before and after each operation to generate textual representations of the action results. This approach captures useful screen information alongside action details while effectively eliminating redundant data. We maintain the most recent action results as \textbf{Short-term Memory(STM)}.We define STM at step $t$ denoted as $M_{t}$, as a set of ordered action-result: $M_{t}=\{r_{1},r_{2},…,r_{k}\}$, where $r_{i}$ is the action result.When a new action 
is executed at step $t+1$, and the action result is summarized as $r_{t+1}$, the STM is updated as follows:
\begin{equation}
M_{t+1}=M_{t}\cup r_{i+1}.
\end{equation}

To ensure that STM reflects only the most recent and relevant task state, it is constrained to hold at most $N$ pairs (where $N$ represents a predefined memory capacity, which is set to 4 in this paper).If $|M_{t+1}|>N$, the oldest memory is removed:
\begin{equation}
M_{t+1}=\{r_{i}|i\in (t-N+2,…,t+1)\}.
\end{equation}

\subsection{Long-term Memory}
Long-term memory facilitates information sharing across applications or tasks, while short-term memory helps GUI Agents maintain a clear understanding of the current task state. Typically, more recent memories are more relevant to immediate decisions, whereas distant short-term memories can become redundant, diluting the effective information density. Conversely, overly brief memory spans risk losing valuable information from previous tasks.

To address this challenge, we designed a long-term memory mechanism. This mechanism begins by extracting task-relevant information $I$ from previous screens, followed by allowing the Agent to distill the core information $C$ from these extractions. Then evaluates whether the distilled information warrants inclusion in long-term memory and make a label $L$ based on its relevance and utility for future tasks. Once deemed valuable, the information is stored and continuously updated in the long-term memory module $M_{LT}$. For step $t$, the process as follows:
\begin{equation}
(L_{t},C_{t}) = f(Q,STM_{t},I_{t}),
\end{equation}
where $Q$ is task Query, $STM$ is short-term memory.

Only information satisfying the condition $C_{LT}=\{C_{t}|L_{t}=true\}$ is stored in the long-term memory module $M_{LT}$. This module is updated according to predefined rules using the function $f_{update}$:
\begin{equation}
M_{LT}\leftarrow f_{update}(M_{LT},C_{LT}).
\end{equation}

This approach ensures that long-term memory retains valuable information while minimizing redundancy, enabling more accurate and efficient decision-making.

\begin{algorithm}[tb]
    \caption{Workflow of Chain-of-Memory}
    \label{alg:algorithm}
    \textbf{Input}: Task Query $\mathit{Q}$, Current Screen $\mathit{S_t}$, Previous Screen $\mathit{S_{t-1}}$, Short-term Memory $\mathit{STM_t}$, Long-term Memory $\mathit{LTM_t}$, Previous Action $\mathit{A_{t-1}}$\\
    \textbf{Parameter}: Maximum STM capacity $N$, Current Step $t$\\
    \textbf{Output}: Current Action $\mathit{A_t}$, Updated $\mathit{STM_{t+1}}$, Updated $\mathit{LTM_{t+1}}$
    \begin{algorithmic}[1] %[1] enables line numbers
        \STATE \textbf{Detect Target:}
        \IF {$a_{t-1} =$ Click}
        \STATE Click Target $\mathit{T_{click}} = DetectTarget(\mathit{S_t}, \mathit{a_{t-1}})$.
        \ELSE
        \STATE $\mathit{T_{click}} = None$.
        \ENDIF
        \STATE \textbf{Short-term Memory Update:}
        \STATE /* $\mathit{AR_t}$: Action Result */
        \STATE $\mathit{AR_t} = Compare(\mathit{T_{click}}, \mathit{A_{t-1}}, \mathit{S_t}, \mathit{S_{t-1}})$
        \STATE $\mathit{STM_t} = \mathit{STM_t} \cup \{\mathit{AR_t}\}$ 
        \IF {$|\mathit{STM_t}| > N$}
        \STATE Remove oldest entry from $\mathit{STM_t}$.
        \ENDIF

        \STATE \textbf{Screen Information Extraction:}
        \STATE $\mathit{ScreenInfo} = ExtractRelevantInfo(\mathit{Q}, \mathit{S_t}, \mathit{STM_t})$

        \STATE \textbf{Long-term Memory Update:}
        \STATE $\mathit{App_t} = DetectApp(\mathit{S_t})$ 
        \STATE $(\mathit{L_{t}}, \mathit{C_t}) = Evaluate(\mathit{Q}, \mathit{STM_t}, \mathit{ScreenInfo})$ 
        \IF {$\mathit{L_t} = \text{true}$}
            \IF {$\mathit{App_t} = \mathit{App_{t-1}}$}
                \STATE Replace the last entry in $\mathit{LTM_t}$ with $\mathit{C_t}$.
            \ELSE
                \STATE Add $\mathit{C_t}$ to $\mathit{LTM_t}$.
            \ENDIF
            \STATE Add $\mathit{App_t}$ to $\mathit{LTM_t}$.
        \ENDIF
        
        \STATE \textbf{Agent Decision:}
        \STATE $\mathit{A_t} = AgentDecision(\mathit{Q}, \mathit{S_t}, \mathit{STM_t}, \mathit{LTM_t})$ 
        
        \STATE \textbf{return} $\mathit{A_t}$, $\mathit{STM_t}$, $\mathit{LTM_t}$
    \end{algorithmic}
\end{algorithm}

\subsection{Screen Information}
Screen information acts as intermediate data that supports the long-term memory module. Directly extracting valuable long-term information from screen captures is challenging for the Agent. However, our experiments show that breaking down tasks into sequential steps enables the Agent to perform significantly better. To ensure the accurate and comprehensive extraction of core information for long-term memory, the Agent identifies the current task goal by referencing the given instruction and short-term memory during the screen information extraction process. It then provides a detailed description exclusively for task-relevant information.

For example, if the current subtask involves searching for tweets related to laptops, the Agent omits descriptions of unrelated screen components, retaining only the full content of tweets relevant to the task. This stepwise approach ensures that the extracted screen information remains both concise and highly relevant to the task objectives.

\begin{figure*}[t]
\vspace{-0.3cm}
\setlength{\belowcaptionskip}{0cm}
\setlength{\abovecaptionskip}{0.1cm}
    \centering
    \includegraphics[width=\linewidth]{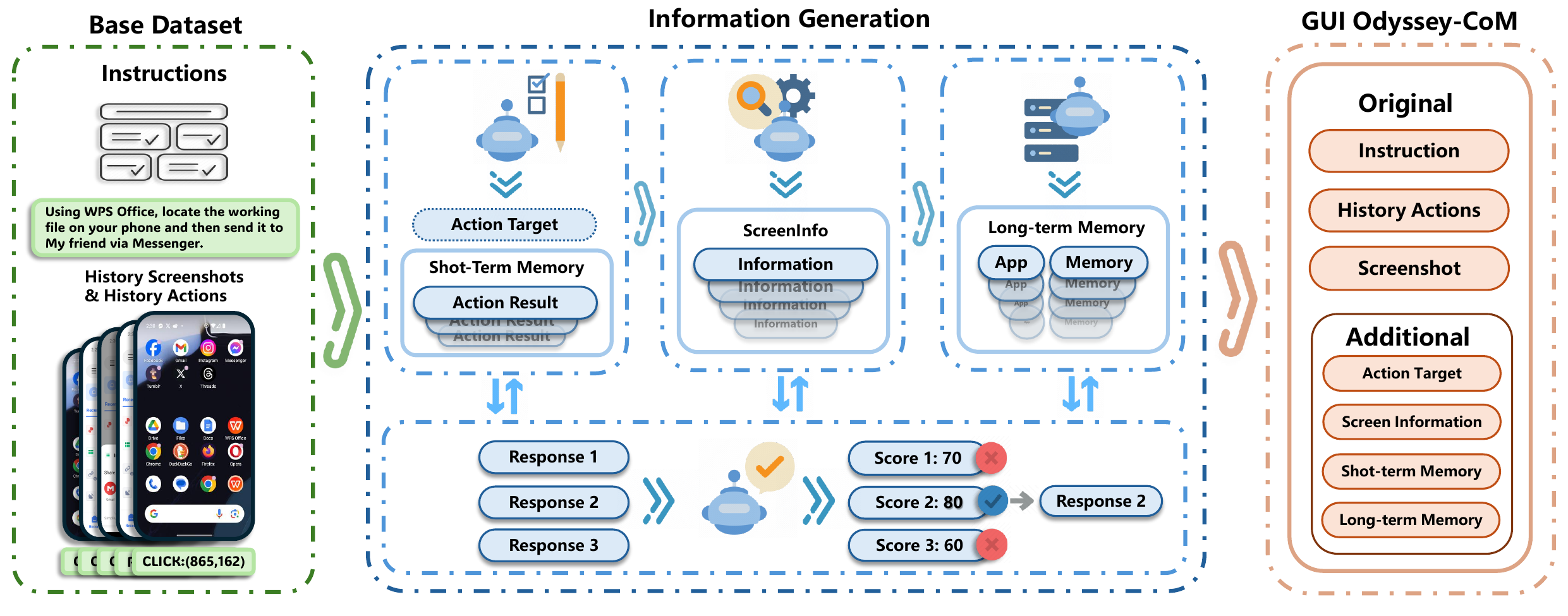}
    \caption{\textbf{GUI Odyssey-CoM collection pipeline. }}
    \label{fig:datapipeline}
    \vspace{-0.2cm}
\end{figure*}

\subsection{Action Decision}
During the Action Decision phase, the agent determines the next operation to be performed on the screen. At this stage, the agent retrieves and integrates task-relevant information from short-term memory and long-term memory with data from the current screen to construct a coherent context. By synthesizing this contextualized information, the agent evaluates possible actions, ensuring that the chosen operation is both contextually appropriate and strategically sound. Following the action, STM and LTM are updated based on the results, establishing a feedback loop that supports continuous improvement.

\begin{table*}[]
    \vspace{0.4cm}
    \setlength{\abovecaptionskip}{0.15cm}
    \setlength{\belowcaptionskip}{0cm}
    \centering
    \resizebox{180mm}{!}{
    %\resizebox{\linewidth}{!}{
    \begin{tabular}{@{}c|c|c|clclclclcl|ccc@{}}
    \toprule
    \multirow{2}{*}{\textbf{Model}} & \multirow{2}{*}{\textbf{FT?}} & \multirow{2}{*}{\textbf{\begin{tabular}[c]{@{}c@{}}Vision\\ History\end{tabular}}} & \multicolumn{10}{c|}{\textbf{Atomic}}                                                                                                                                                        & \multicolumn{3}{c}{\textbf{Total}}               \\ \cmidrule(l){4-16} 
                                    &                              &                                                                                    & \multicolumn{2}{c|}{\textbf{CLICK}} & \multicolumn{2}{c|}{\textbf{TYPE}}  & \multicolumn{2}{c|}{\textbf{SCROLL}} & \multicolumn{2}{c|}{\textbf{HOME}}  & \multicolumn{2}{c|}{\textbf{STOP}}  & \textbf{AMS}   & \textbf{SR}    & \textbf{TSS}    \\ \midrule
    \textbf{Qwen2-VL-7B}            & $\usym{2717}$                & $\usym{2717}$                                                                      & \multicolumn{2}{c|}{11.1}           & \multicolumn{2}{c|}{36.85}          & \multicolumn{2}{c|}{15.77}           & \multicolumn{2}{c|}{4.66}           & \multicolumn{2}{c|}{25.12}          & 14.6           & 0              & 2.55           \\
    \textbf{+CoM}                   & $\usym{2717}$                & $\usym{2717}$                                                                      & \multicolumn{2}{c|}{8.45}           & \multicolumn{2}{c|}{41.37}          & \multicolumn{2}{c|}{\textbf{21.71}}  & \multicolumn{2}{c|}{3.0}            & \multicolumn{2}{c|}{8.37}           & 12.71          & 0              & 1.25           \\ \midrule
    \textbf{InternVL2.5-78B}        & $\usym{2717}$                & $\usym{2717}$                                                                      & \multicolumn{2}{c|}{34.63}          & \multicolumn{2}{c|}{60.49}          & \multicolumn{2}{c|}{6.26}           & \multicolumn{2}{c|}{47.34}          & \multicolumn{2}{c|}{8.64}           & 33.72          & 0.31           & 12.67          \\
    \textbf{+CoM}                   & $\usym{2717}$                & $\usym{2717}$                                                                      & \multicolumn{2}{c|}{38.61}          & \multicolumn{2}{c|}{\textbf{67.59}} & \multicolumn{2}{c|}{8.94}            & \multicolumn{2}{c|}{\textbf{58.84}} & \multicolumn{2}{c|}{27.25}          & 39.38          & 0.62           & \textbf{30.36} \\ \midrule
    \textbf{Qwen2-VL-72B}           & $\usym{2717}$                & $\usym{2717}$                                                                      & \multicolumn{2}{c|}{35.39}          & \multicolumn{2}{c|}{36.33}          & \multicolumn{2}{c|}{16.53}           & \multicolumn{2}{c|}{31.08}          & \multicolumn{2}{c|}{40.69}          & 33.44          & 0.36           & 6.27           \\
    \textbf{+CoM}                   & $\usym{2717}$                & $\usym{2717}$                                                                      & \multicolumn{2}{c|}{\textbf{40.28}} & \multicolumn{2}{c|}{54.16}          & \multicolumn{2}{c|}{17.08}           & \multicolumn{2}{c|}{34.3}           & \multicolumn{2}{c|}{\textbf{61.92}} & \textbf{40.18} & \textbf{0.67}  & 18.32          \\ \midrule \midrule
    \textbf{OdysseyAgent}           & $\usym{2713}$                & $\usym{2713}$                                                                      & \multicolumn{2}{c|}{78.72}          & \multicolumn{2}{c|}{82.59}          & \multicolumn{2}{c|}{\textbf{77.97}}  & \multicolumn{2}{c|}{83.3}           & \multicolumn{2}{c|}{94.13}          & 79.99          & 14.43          & 73.0           \\ \midrule
    \textbf{Qwen-VL-7b}             & $\usym{2713}$                & $\usym{2717}$                                                                      & \multicolumn{2}{c|}{80.2}           & \multicolumn{2}{c|}{82.04}          & \multicolumn{2}{c|}{70.3}            & \multicolumn{2}{c|}{82.36}          & \multicolumn{2}{c|}{91.73}          & 79.98          & 15.0           & 72.01          \\
    \textbf{+CoM}                   & $\usym{2713}$                & $\usym{2717}$                                                                      & \multicolumn{2}{c|}{81.62}          & \multicolumn{2}{c|}{87.21}          & \multicolumn{2}{c|}{69.25}           & \multicolumn{2}{c|}{85.62}          & \multicolumn{2}{c|}{94.03}          & 81.74          & 18.0           & 79.4           \\ \midrule
    \textbf{Qwen2-VL-7b}            & $\usym{2713}$                & $\usym{2717}$                                                                      & \multicolumn{2}{c|}{83.54}          & \multicolumn{2}{c|}{82.2}           & \multicolumn{2}{c|}{72.4}            & \multicolumn{2}{c|}{86.21}          & \multicolumn{2}{c|}{94.03}          & 82.84          & 19.14          & 75.68          \\
    \textbf{+CoM}                   & $\usym{2713}$                & $\usym{2717}$                                                                      & \multicolumn{2}{c|}{84.4}           & \multicolumn{2}{c|}{\textbf{88.15}} & \multicolumn{2}{c|}{71.56}           & \multicolumn{2}{c|}{\textbf{88.49}} & \multicolumn{2}{c|}{\textbf{96.16}} & \textbf{84.25} & 21.0           & \textbf{83.97} \\
    \textbf{+CoM*}                  & $\usym{2713}$                & $\usym{2717}$                                                                      & \multicolumn{2}{c|}{\textbf{84.51}} & \multicolumn{2}{c|}{86.41}          & \multicolumn{2}{c|}{74.78}           & \multicolumn{2}{c|}{85.31}          & \multicolumn{2}{c|}{95.89}          & 84.22          & \textbf{21.16} & 81.33     
    \\ \bottomrule
    \end{tabular}}
    
    \caption{\textbf{Main results of Chain-of-Memory.} FT indicates whether fine-tuning has been applied, corresponding to zero-shot evaluation and fine-tuning evaluation. Vision History refers to whether images are used as historical information. * Indicates that all memory information is autonomously generated by the agent. Among these sub-operations, CLICK and TYPE involve coordinates and text, making it challenging to achieve a perfect match with the ground truth. Therefore, alternative calculation methods are introduced. For the remaining operations, the evaluation determines whether they fully match the ground truth.}
    \label{tab:main1}
    \end{table*}

\section{GUI Odyssey-CoM}  
%-------------------------------------------------------------------------

\subsection{Data Collection}
Existing datasets lack CoM annotations, which hinders smaller models from acquiring memory management capabilities. To address this, we constructed a cross-app dataset annotated with CoM and designed a data generation pipeline to ensure high data quality.We constructed our dataset based on GUI Odyssey, the largest publicly available cross-app GUI navigation dataset. GUI Odyssey comprises 7735 episodes, each with unique instructions. It also encompasses a diverse range of device types, app categories, and task types.

%-------------------------------------------------------------------------
\subsection{Pretrained Model Selection}
Recently, dozens of MLLMs have been released in the open-source community. Based on their evaluation metrics, we have initially selected three models: Qwen2-VL-72B~\cite{wang2024qwen2}, InternVL2.5-78B~\cite{wang2024mpo,chen2024expanding}, and Llama-3.2-90B-Vision~\cite{llamaV}. To further compare the capabilities of these models in recognizing and annotating GUI navigation datasets, we randomly sampled 50 episodes from the GUI-Odyssey dataset and then manually selected 20 episodes featuring diverse tasks and applications.

We conducted a manual evaluation of the models' performance in generating CoM data. Based on this evaluation, Qwen2-VL-72B was ultimately chosen as the preferred model.

%-------------------------------------------------------------------------
\subsection{Information Generation}
The Chain-of-Memory (CoM) framework processes information in a specific order: short-term memory, screen information, and long-term memory. These stages are interdependent, with earlier outputs serving as inputs for later stages. To ensure alignment with this framework, our data generation process follows the same order. Annotations are completed for earlier stages first and then used to generate annotations for subsequent stages. The overall workflow is illustrated in Figure \ref{fig:datapipeline}.

To maintain data quality, we developed a robust evaluation mechanism. The data generation model was configured to produce three responses for each input. Three evaluation standards, tailored to each data type, were defined. An evaluation model scored the three responses for each input based on these standards. The response with the highest score was selected as the final annotation and stored.

The model used for data generation in each step strictly adheres to the CoM framework described earlier. This consistency ensures the process's reliability and prevents any potential data leakage.

\begin{table*}[htb]
\vspace{-0.4cm}
\setlength{\abovecaptionskip}{0.15cm}
\setlength{\belowcaptionskip}{0cm}
\centering
\resizebox{180mm}{!}{
\begin{tabular}{@{}c|ccc|cclclcccl|cc@{}}
\toprule
\multirow{2}{*}{\textbf{MODE}} & \multicolumn{3}{c|}{\textbf{Semantic Annotations}}             & \multicolumn{9}{c|}{\textbf{Atomic}}                                                                                                                                                   & \multicolumn{2}{c}{\textbf{Episodic}}               \\ \cmidrule(l){2-15} 
                               & \textbf{ScreenInfo} & \textbf{Short-term} & \textbf{Long-term} & \textbf{CLICK} & \multicolumn{2}{c}{\textbf{TYPE}}  & \multicolumn{2}{c}{\textbf{SCROLL}} & \textbf{HOME}  & \multicolumn{1}{c|}{\textbf{STOP}}  & \multicolumn{2}{c|}{\textbf{AMS}}   & \multicolumn{1}{c|}{\textbf{SR}}   & \textbf{TSS}   \\ \midrule
\multirow{6}{*}{\textbf{ZS}}   &                     &                     &                    & 35.39          & \multicolumn{2}{c}{36.33}          & \multicolumn{2}{c}{16.53}           & 31.08          & \multicolumn{1}{c|}{40.69}          & \multicolumn{2}{c|}{33,44}          & \multicolumn{1}{c|}{0.36}          & 6.27           \\ \cmidrule(l){2-15} 
                               & $\checkmark$        &                     &                    & 32.55          & \multicolumn{2}{c}{38.9}           & \multicolumn{2}{c}{7.49}            & 26.6           & \multicolumn{1}{c|}{58.93}          & \multicolumn{2}{c|}{31.85}          & \multicolumn{1}{c|}{0.31}          & 7.08           \\
                               &                     & $\checkmark$        &                    & 39.05          & \multicolumn{2}{c}{47.8}           & \multicolumn{2}{c}{17.0}            & \textbf{37.93} & \multicolumn{1}{c|}{51.15}          & \multicolumn{2}{c|}{38.28}          & \multicolumn{1}{c|}{\textbf{0.72}} & 18.18          \\
                               &                     &                     & $\checkmark$       & 34.45          & \multicolumn{2}{c}{44.72}          & \multicolumn{2}{c}{11.69}           & 27.5           & \multicolumn{1}{c|}{43.36}          & \multicolumn{2}{c|}{33.18}          & \multicolumn{1}{c|}{0.36}          & 6.27           \\ \cmidrule(l){2-15} 
                               &                     & $\checkmark$        & $\checkmark$       & \textbf{40.28} & \multicolumn{2}{c}{\textbf{54.16}} & \multicolumn{2}{c}{\textbf{17.08}}  & 34.3           & \multicolumn{1}{c|}{61.92}          & \multicolumn{2}{c|}{\textbf{40.18}} & \multicolumn{1}{c|}{0.67}          & \textbf{18.32} \\ \cmidrule(l){2-15} 
                               & $\checkmark$        & $\checkmark$        & $\checkmark$       & 38.72          & \multicolumn{2}{c}{45.2}           & \multicolumn{2}{c}{11.4}            & 24.05          & \multicolumn{1}{c|}{\textbf{67.04}} & \multicolumn{2}{c|}{37.22}          & \multicolumn{1}{c|}{0.62}          & 13.43          \\ \midrule \midrule
\multirow{6}{*}{\textbf{FT}}   &                     &                     &                    & 83.54          & \multicolumn{2}{c}{82.2}           & \multicolumn{2}{c}{72.4}            & 86.21          & \multicolumn{1}{c|}{94.03}          & \multicolumn{2}{c|}{82.84}          & \multicolumn{1}{c|}{19.14}         & 75.68          \\ \cmidrule(l){2-15} 
                               & $\checkmark$        &                     &                    & 83.43          & \multicolumn{2}{c}{81.82}          & \multicolumn{2}{c}{71.74}           & 83.3           & \multicolumn{1}{c|}{93.97}          & \multicolumn{2}{c|}{82.42}          & \multicolumn{1}{c|}{17.74}         & 75.15          \\
                               &                     & $\checkmark$        &                    & 83.63          & \multicolumn{2}{c}{83.62}          & \multicolumn{2}{c}{\textbf{75.43}}  & 86.83          & \multicolumn{1}{c|}{95.47}          & \multicolumn{2}{c|}{83.5}           & \multicolumn{1}{c|}{19.5}          & 82.04          \\
                               &                     &                     & $\checkmark$       & 84.24          & \multicolumn{2}{c}{87.57}          & \multicolumn{2}{c}{74.78}           & 85.49          & \multicolumn{1}{c|}{93.49}          & \multicolumn{2}{c|}{84.0}           & \multicolumn{1}{c|}{20.95}         & 78.55          \\ \cmidrule(l){2-15} 
                               &                     & $\checkmark$        & $\checkmark$       & \textbf{84.4}  & \multicolumn{2}{c}{\textbf{88.15}} & \multicolumn{2}{c}{71.56}           & \textbf{88.49} & \multicolumn{1}{c|}{\textbf{96.16}} & \multicolumn{2}{c|}{\textbf{84.25}} & \multicolumn{1}{c|}{\textbf{21.0}} & \textbf{83.97} \\ \cmidrule(l){2-15} 
                               & $\checkmark$        & $\checkmark$        & $\checkmark$       & 82.05          & \multicolumn{2}{c}{85.48}          & \multicolumn{2}{c}{71.45}           & 85.27          & \multicolumn{1}{c|}{95.41}          & \multicolumn{2}{c|}{82.11}          & \multicolumn{1}{c|}{16.3}          & 79.49          \\ \bottomrule
\end{tabular}}
\caption{\textbf{Ablation study of CoM components on Qwen2-VL-72B/7B}. This ablation study includes both zero-shot and fine-tuning experiments. The first row of each experiment represents the results without any additional information, serving as the baseline. The remaining rows present the results obtained after incorporating the corresponding additional information.}
\label{tab:Ablation}
\end{table*}

\section{Experiments}
In this section, we conduct zero-shot evaluation and fine-tuning evaluation to assess the effectiveness of the CoM in enhancing GUI Agent decision-making. Additionally, to evaluate whether GUI Odyssey-CoM can enable smaller models to generate and manage memory effectively, we trained a model based on this dataset that possesses both memory generation and decision-making capabilities. Furthermore, we performed ablation studies to investigate the individual contributions of long-term and short-term memory to the performance of the GUI Agent.

\subsection{Experiment Setup}
\subsubsection{Zero-Shot Evaluation} To enable the Agent to fully comprehend and utilize the information stored in memory, we selected the state-of-the-art open-source multimodal large language model, Qwen2-VL-72B~\cite{wang2024qwen2}, for the zero-shot evaluation. This model incorporates a native dynamic resolution mechanism, allowing it to align with human perception processes. Furthermore, it exhibits strong agent capabilities, outperforming GPT-4o on several agent benchmarks.We conducted zero-shot evaluations of the Qwen2-VL-72B model on both the original GUI Odyssey dataset and GUI Odyssey-CoM to validate the performance improvements of CoM in GUI agent tasks.
\subsubsection{Fine-Tuning Evaluation} The fine-tuning evaluation was conducted using the OdysseyAgent~\cite{lu2024gui}, its base model, Qwen-VL-7B, and the 7B version of Qwen2-VL. Models were fine-tuned and evaluated sequentially on both the GUI Odyssey and GUI Odyssey-CoM to validate whether CoM data can execute user instructions effectively and better assist the model in learning how to generate memory information.

%-------------------------------------------------------------------------
\subsubsection{Evaluation Metrics}
We adopt \textbf{Action Matching Score (AMS)} and \textbf{Success Rate (SR)} as primary evaluation metrics, following \cite{lu2024gui} and \cite{zhang2024android}. \textbf{AMS} quantifies the accuracy of individual actions, while \textbf{SR} measures the percentage of successfully completed episodes. The evaluation criteria for each action are consistent with those used in GUI Odyssey \cite{lu2024gui}. Furthermore, we introduce \textbf{Task Switching Score (TSS)}, a metric specifically designed to measure the agent's ability to correctly switch between subtasks. \textbf{TSS} calculates the success rate of task switching. If the ground truth of the current step is HOME, then if both the current step and the next step are correct, it is considered success; otherwise, it is failure. 
%TSS计算Agent的任务切换成功率，如果当前step的ground truth是HOME，那么如果当前step和下一step都正确则认为任务切换成功，反之则不成功。

%Additionally, we introduce a new metric, \textbf{Task Switching Score (TSS)}, which measures the success rate of subtask switching in cross-app tasks, reflecting the Agent's understanding of both the current task state and task goals. The calculation method is as follows: if the current ground truth operation is HOME, and both the current and next operations are correct, the task switch is deemed successful. The evaluation criteria for each action are consistent with those used in GUI-Odyssey. Specifically, for CLICK actions, we evaluate whether the distance to the gold gestures is below a predefined threshold. For TYPE actions, we assess whether the Average Normalized Levenshtein Similarity (ANLS) with the gold gestures exceeds a predefined threshold. The thresholds are set at 14\% for CLICK and 50\% for TYPE \cite{lu2024gui}.

\subsubsection{Implementation details}
GUI Odyssey-CoM extends GUI Odyssey by incorporating ScreenInfo, short-term memory, and long-term memory. For \textbf{Zero-shot Evaluation}, we utilize the test sets of both datasets. Following \cite{zhang2024android}, we tailor prompts for each experiment group to ensure proper task understanding. Our baseline is established using the model's outputs without any additional information, and we incrementally add information to assess the impact of each component on the model's performance. To ensure experimental efficiency and stability, we adopt AWQ-quantized \cite{lin2024awq} models with greedy decoding. For \textbf{Fine-Tuning Evaluation}, we employ two training methods: \textbf{(1) Conventional Fine-tuning:} We train the model on datasets with various combinations of additional information and evaluate its performance. The model trained without any additional information serves as the baseline in this setting. \textbf{(2) Memory-Enhanced Fine-tuning:} We directly enhance the model's memory generation and utilization capabilities by training it with the long-term and short-term memories from the GUI Odyssey-CoM training set. This method allows the model to learn to generate accurate memory information and effectively leverage it during decision-making.

%-------------------------------------------------------------------------

\subsection{Comparative Result}
Table \ref{tab:main1} demonstrates that utilizing CoM, with or without fine-tuning, significantly enhances the performance of the GUI Agent. Long-term memory proves particularly beneficial for text-based operations, such as retrieving the price of a previously searched laptop with favorable reviews or sharing a previously identified queue address. The accuracy of \textbf{TYPE} actions improves significantly because these operations rely on textual input to access long-term memory.

In contrast, actions like \textbf{CLICK}, \textbf{HOME}, and \textbf{STOP} depend more on understanding current task states and overall goals. For example, pressing \textbf{Home} to switch task after completing a subtask or executing \textbf{COMPLETE} to terminate the workflow after completing all subtasks. Short-term memory enhances the agent's ability to track current task state, leading to significant improvements in the accuracy of these actions.

In zero-shot experiment, the 7B model initially struggled with CoM, likely due to its smaller scale, which limits its ability to directly process CoM information. However, after fine-tuning, the 7B model exhibited excellent performance, indicating that the GUI Odyssey-CoM dataset provided the necessary training for effective CoM utilization. Notably, the 7B model's self-generated memory information proved nearly as effective in decision-making as that generated by the 72B model, further emphasizing the importance of the dataset. 

\begin{figure}
\setlength{\belowcaptionskip}{0cm}
\setlength{\abovecaptionskip}{0cm}
    \centering
    \includegraphics[width=\linewidth]{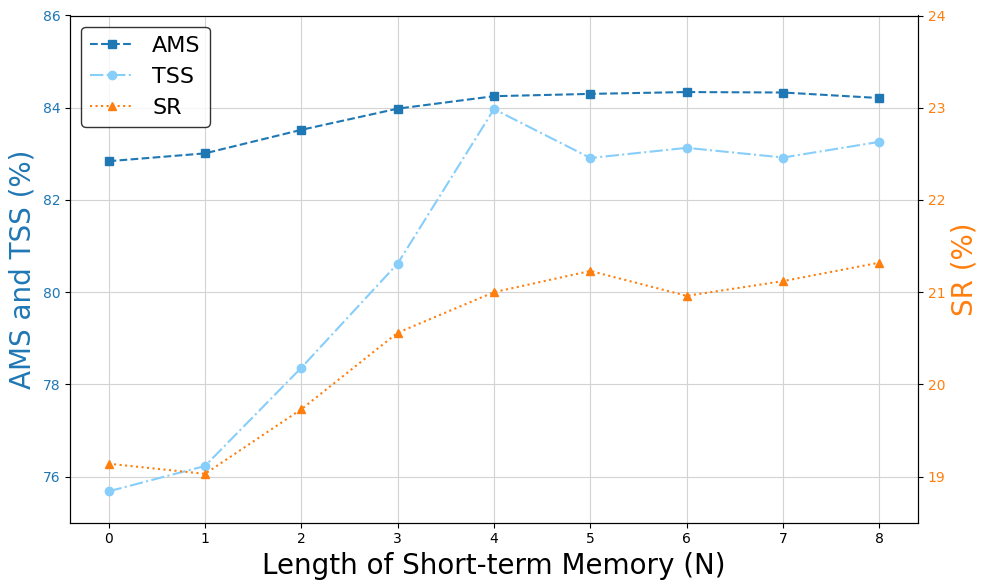}
    \caption{Influence of short-term memory length.}
    \label{fig:Ablation2}
    \vspace{0.0cm}
\end{figure}

\subsection{Ablation Study}
To evaluate the impact of each CoM component on the performance of GUI agents, we conducted ablation studies using the Qwen2-VL-72B model for zero-shot experiments and the Qwen2-VL-7B model for fine-tuning experiments.

\paragraph{\textbf{Component Contribution}} Table \ref{tab:Ablation} shows that \textbf{Short-term memory} significantly improves the accuracy of \textbf{HOME} and \textbf{STOP} actions compared to the baseline, while \textbf{Long-term memory} notably improves the accuracy of text input \textbf{(TYPE)} actions. Combining both short-term and long-term memory reached the highest overall AMS, SR and TSS, indicating that short-term memory enhances the agent's ability to track immediate task progress and long-term memory primarily facilitates information transfer between tasks. However, whether in zero-shot or fine-tuning experiments, when \textbf{ScreenInfo} is used as input, it will significantly reduce the performance of the GUI Agent.This phenomenon may be attributed to the potential for extraneous information present on certain screens to impede decision-making, a limitation that persists even when ScreenInfo is integrated with other memory components. These findings underscore the necessity of selective information retention within the memory architecture. In aggregate, the results corroborate the efficacy of the CoM framework, particularly the synergistic interplay between long-term and short-term memory, in navigating the complexities of cross-application tasks.

\paragraph{\textbf{The influence of short-term memory length}} Figure \ref{fig:Ablation2} illustrates the impact of short-term memory length $N$ on the GUI Agent's performance, measured by AMS, TSS and SR. \textbf{AMS} exhibits a positive correlation with N, suggesting that longer short-term memory, within a certain range, improves overall action accuracy. However, this trend plateaus beyond N=4, indicating diminishing returns with increasing memory length. Similar patterns are observed for \textbf{TSS} and \textbf{SR}. In conclusion, increasing short-term memory length positively impacts agent performance within a specific range, but excessively long short-term memory provides minimal additional benefit, potentially due to utilization limitations or increased computational overhead.

\section{Conclusion}
Existing GUI Agents struggle with perceiving task states and retaining critical information in cross-app tasks, which limits their ability to complete tasks. To address this, we introduced the CoM paradigm and constructed a high-quality CoM dataset, GUI Odyssey-CoM. The CoM paradigm enables GUI Agents to overcome these challenges by managing both long-term and short-term memory, while the GUI Odyssey-CoM allows smaller models to achieve memory generation capabilities comparable to those of larger models. Experimental results demonstrate the effectiveness of the proposed Chain-of-Memory approach. Furthermore, the GUI Odyssey-CoM contributes to the future enhancement of memory capabilities in GUI Agents.

\appendix

\bibliographystyle{named}
\bibliography{ref}

\newpage
\newpage

\end{document}